\pdfoutput=1

\documentclass[11pt]{article}

\usepackage[]{acl}

\usepackage{times}
\usepackage{latexsym}
\usepackage{graphicx}
\usepackage{subfigure}
\usepackage{hyperref}       
\usepackage{algorithmic}
\usepackage{algorithm}
\usepackage{tabularx}
\usepackage{amsmath}

\usepackage[T1]{fontenc}

\usepackage[utf8]{inputenc}

\usepackage{microtype}

\usepackage{inconsolata}

\usepackage{array}
\usepackage{pifont}
\usepackage{tabularx}
\usepackage{adjustbox}
\usepackage{multirow}
\usepackage{enumitem}
\usepackage{xspace}
\usepackage{tcolorbox}
\usepackage{booktabs,amsfonts,dcolumn}
\usepackage{hyperref}
\usepackage{url}
\usepackage{amsmath,amsthm,amsfonts,amssymb,bm,stmaryrd,bbm}
\usepackage[noorphans,vskip=0.75ex,leftmargin=2ex]{quoting}
\usepackage{footmisc}\interfootnotelinepenalty=10000
\usepackage{colortbl}
\usepackage[noabbrev,capitalize]{cleveref}

\newtcolorbox[list inside=prompt,auto counter,number within=section]{prompt}[1][]{
    colbacktitle=black!60,
    coltitle=white,
    fontupper=\footnotesize,
    boxsep=5pt,
    left=0pt,
    right=0pt,
    top=0pt,
    bottom=0pt,
    boxrule=1pt,
    #1,
}

\renewcommand{\paragraph}[1]{\vspace{0.2cm}\noindent\textbf{#1}}

\newcommand{\ours}{{\sc{QRMeM}}}
\newcommand{\TRIAL}{{\sc{Entity Trial}}}
\newcommand{\NAVI}{{\sc{Knowledge-directed Navigation}}}
\newcommand{\RETRIEVAL}{{\sc{Graph Expansion Search}}}
\newcommand\blue[1]{\textbf{\color{blue}{#1}}}

\usepackage{amsmath,amsfonts,bm}

\def\eqref#1{equation~\ref{#1}}

\def\1{\bm{1}}

\def\rmE{{\mathbf{E}}}

\def\rmR{{\mathbf{R}}}
\def\rmS{{\mathbf{S}}}

\def\ermF{{\textnormal{F}}}

\def\ermK{{\textnormal{K}}}

\DeclareMathAlphabet{\mathsfit}{\encodingdefault}{\sfdefault}{m}{sl}
\SetMathAlphabet{\mathsfit}{bold}{\encodingdefault}{\sfdefault}{bx}{n}

\def\sM{{\mathbb{M}}}
\def\sN{{\mathbb{N}}}

\def\sQ{{\mathbb{Q}}}

\def\sS{{\mathbb{S}}}

\title{QRMeM: Unleash the Length Limitation through Question then Reflection Memory Mechanism}
\author{Bo Wang\textsuperscript{1,3} , Heyan Huang\textsuperscript{1,3}, Yixin Cao\textsuperscript{2}, Jiahao Ying\textsuperscript{4},
     {\bf Wei Tang\textsuperscript{5}},  {\bf Chong Feng\textsuperscript{1,3} } \\       
        \textsuperscript{1}Beijing Institute of Technology, China
        \\ \textsuperscript{2}Fudan University, China
        \\
        \textsuperscript{3}Southeast Academy of Information Technology, Beijing Institute of Technology, China
        \\
        \textsuperscript{4}Singapore Management University, Singapore
        \\ \textsuperscript{5} University of Science and Technology of China, China
        \\ \texttt{bwang@bit.edu.cn}
        }

\begin{document}
\maketitle
\begin{abstract}
While large language models~(LLMs) have made notable advancements in natural language processing, they continue to struggle with processing extensive text. 
Memory mechanism offers a flexible solution for managing long contexts, utilizing techniques such as compression, summarization, and structuring to facilitate nuanced and efficient handling of large volumes of text. 
However, existing techniques face challenges with static knowledge integration, leading to insufficient adaptation to task-specific needs and missing multi-segmentation relationships, which hinders the dynamic reorganization and logical combination of relevant segments during the response process.
To address these issues, we introduce a novel strategy, Question then Reflection Memory Mechanism (\ours{}), incorporating a dual-structured memory pool. 
This pool synergizes static textual content with structured graph guidance, fostering a reflective trial-and-error approach for navigating and identifying relevant segments. 
Our evaluation across multiple-choice questions~(MCQ) and multi-document question answering~(Multi-doc QA) benchmarks showcases \ours{}’s enhanced performance compared to existing approaches. Our code is available at~\url{https://github.com/wang202111/qrmem}.

\end{abstract}

\section{Introduction}
Large language models~(LLMs) have achieved significant advancements in natural language processing. Nevertheless, they encounter difficulties handling extensive information, underscoring the necessity for efficient long-context processing.
Expanding the context window size is a crucial way to handle the problem, including positional embedding and key-value cache~\cite{beltagy2020longformer, zaheer2020bigbird,chen2023extending}. 
However, the inclusion of more information inevitably results in the accumulation of irrelevant document content.

Retrieval-based methods aim to filter relevant parts from the original long documents~\cite{zhang2023merging, yu2023augmentation}. They achieve this through various retrieval operations, such as dense, sparse, and reranking, between the text and the query.
\begin{figure}[t]
    \centering
\includegraphics[width=0.45\textwidth]{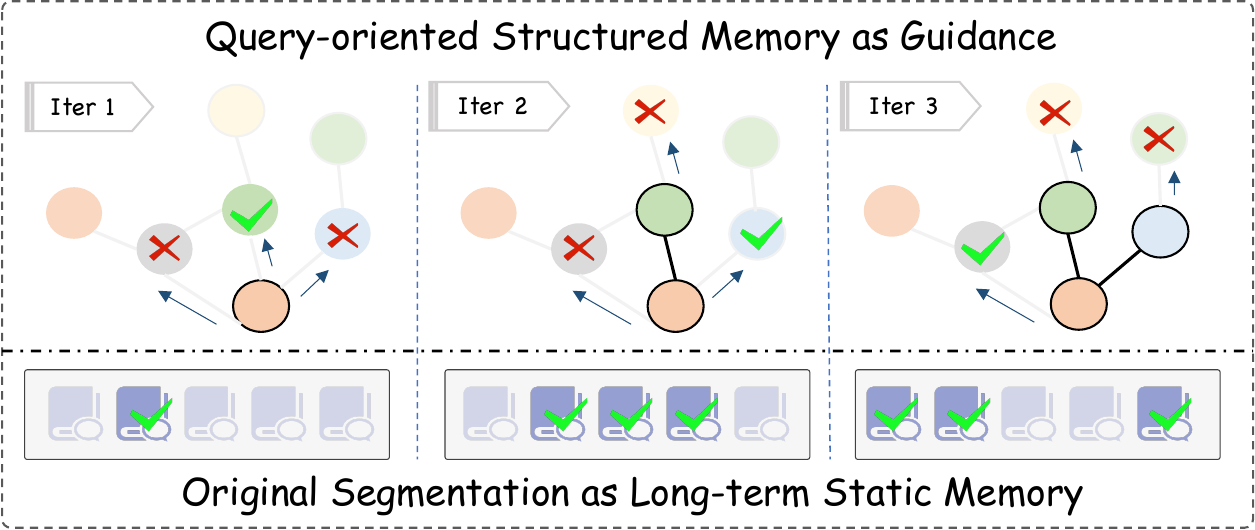}
    \caption{Question then Reflection Memory mechanism is divided into two parts. The upper part, Structured Memory, is a graph constructed from entities and relations extracted from the text. The lower part, Original Segmentation, is the original segmentation. These two parts are linked by the co-occurrence relationships of entities appearing in the segments. \ours{} starts from the core entity of the question and expands on the Structured Memory to find the most relevant next entity node. 
    It uses this to expand and revise the segment sets until the corresponding segment set can answer the question. 
    }
    \label{fig:simple}
\vspace{-10pt}
\end{figure}
Another line of approaches, known as memory mechanism methods~\cite{scm2023liang}, takes the source as a memory pool. These methods offer a solution for managing long contexts and provide an opportunity to handle extraneous documents effectively.
Existing memory mechanisms employ several strategies for indexing extensive information. The strategies include storing most recentest acquired information~\cite{scm2023liang,ebbinghaus2013memory}, selecting contents based on their relevance, importance, and topics~\cite{mmbankZhongGGYW24,gaParkOCMLB23}, and handling long contexts through techniques such as compression~\cite{pan2024llmlingua2}, summarization~\cite{chen2023walking}, and structuring~\cite{chatdb2023hu}. 

Despite the existing strategies, several critical issues still persist due to limitations in the accuracy of retrieval modules or other memory construct tools.
First, the memory mechanism is built primarily on human prior knowledge or predefined structure, which may not always align perfectly with the nuanced needs of real-world applications. This results in the gap between the practical scenario queries and the predefined indexing that occurs in the memory mechanism, making it challenging to find relevant information.
Additionally, tools or LLMs could be employed to manage extensive information in the memory through processes such as compression, structuring, or other methods, which can lead to error propagation or information loss. Consequently, this can result in biased answers based on memory.
Moreover, pinpointing the exact relevant information within these long contexts is inherently challenging. This difficulty is exacerbated by the needs for multi-hop reasoning, long-distance dependencies, and distinguishing between similar segments, which requires tracing information across multiple steps or layers of data, making it difficult to directly locate the segments needed for answering questions.

Socratic inquiry involves asking questions to stimulate critical thinking and illuminate ideas. Inspired by this, we propose a Question then Reflection based Memory Mechanism, which guides thinking through questioning and directs the next steps through reflection on the current state, shown in Figure~\ref{fig:simple}. This mechanism enables the utilization of structured knowledge as guidance to mine the supporting segment.
To mitigate the gap between practical scenario queries and predefined indexing, the query-oriented structured memory is constructed by dynamically aligning with the query’s needs. The structured memory is formed as a graph with entities and open relations as nodes and edges, which could be a guide to mine the supporting segment.
To address information bias during memory construction, we additionally include the original segments as a long-term static memory. These segments are associated with the entity to leverage structured memory for further utilization. Thus, only the filtered segments are used to ensure information fidelity when answering the final question.
As for addressing the challenge of pinpointing dispersed relevant segments, we propose three strategies,
\TRIAL{} achieves comparable results by navigating through only the information of entities and segments in the memory. 
\RETRIEVAL{} uses the relationships between entity nodes as guidance, thereby alleviating complex issues during the retrieval process.
\ours{} reflects on error information from past experiences during multiple iterations, which makes the subsequent steps more effective in finding the relevant segmentation.
To summarize, our main contributions are as follows:
\vspace{-4pt}
\begin{itemize}[leftmargin=*]
    \vspace{-4pt}
    \item A query-oriented method is proposed to bridge the gap between predefined memory and queries while a dual-structure memory pool is built to mitigate the influence of reorganizing memory.
    \vspace{-8pt}
    \item Three distinct methodologies are explored utilizing the memory pool, including \ours{}, \TRIAL{}, and \RETRIEVAL{}. These methodologies adopt different strategies in the interaction process with memory.
    \vspace{-8pt}
    \item  Experimental results on multiple-choice questions~(MCQ) and multi-document question answering~(Multi-doc QA) show that our model achieves superior results with the various methods, validating the effectiveness of our approach.
\end{itemize}
\vspace{-6pt}
\vspace{-6pt}
\section{Methodology}
\vspace{-3pt}
\begin{figure*}[t]
    \centering
    \includegraphics[width=0.9\textwidth]{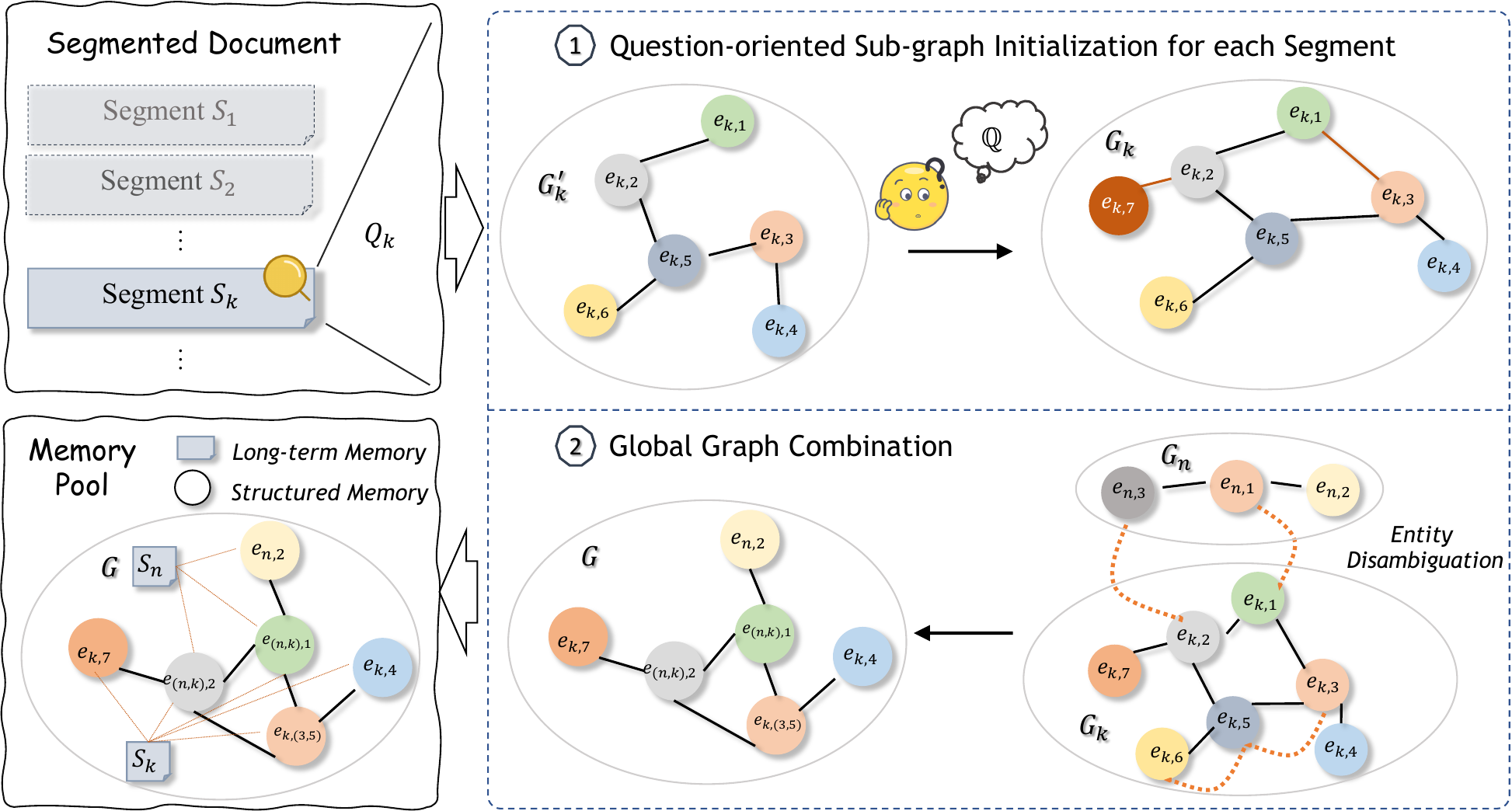}
    \caption{Dual-structure Memory Construction pipeline in \ours{}, where $e$ is the entity inside each graph. \normalsize{\textcircled{\scriptsize{1}}}\normalsize \ For each segment, the sub-graph $G_k$ is initialized according to the question $Q$ to be answered, then updated by the further generated questions $\sQ$ to the current graph. \normalsize{\textcircled{\scriptsize{2}}}\normalsize \ Sub-graphs are combined into global graph $G$ through entity disambiguation and relation fusion. All the segmentation $\rmS_k$ is linked with the entity that appeared to construct the dual-structured memory pool.
    }
    \label{fig:ms}
    \vspace{-10pt}
\end{figure*}

The components of \ours{} are divided into Dual-structure Memory Construction and Reflection-based Relevant Segments Navigation.

\begin{figure*}[t]
    \centering
\includegraphics[width=0.85\textwidth]{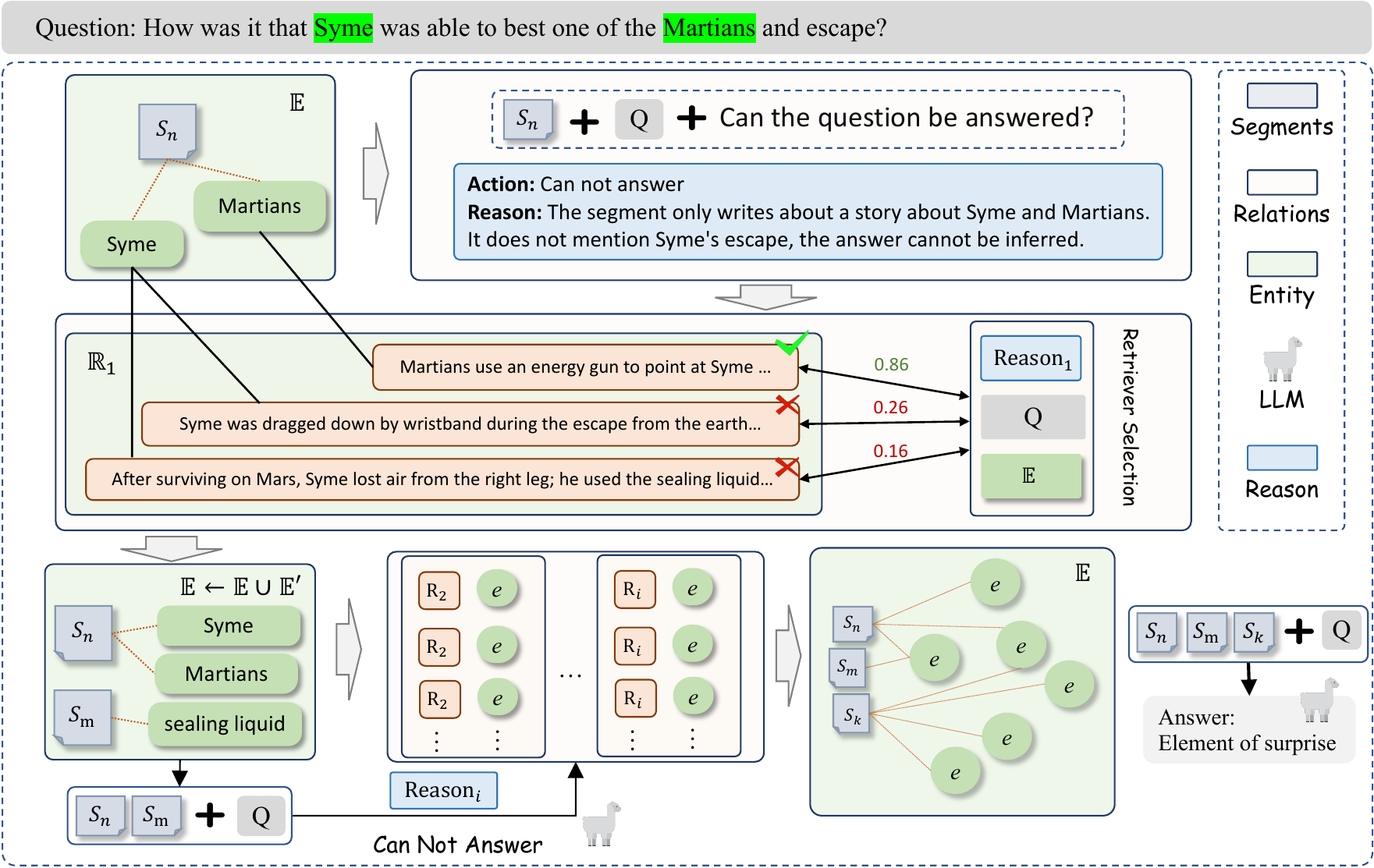}
    \caption{Strategies for \NAVI{} in \ours{}. Navigation starts with a subgraph containing only the target entity from the question and its linked segments. The subgraph is then expanded based on the relevance of relationship edges connected to the current entity. The process will continue until the linked segments are sufficient to answer the question.}
    \label{fig:navi}
\vspace{-10pt}
\end{figure*}

\subsection{Dual-structure Memory Construction}
The approach involves partitioning the memory component into two distinct sections, structured and long-term static memory. 
The construction is depicted in Figure~\ref{fig:ms}.
The structured aspects of the memory resemble human memory anchors, assisting in finding more relevant chunks related to the query as one's understanding deepens.
The Structured Memory serves only as guidance to avoid error accumulation during the graph construction process, while the Original Segmentation is used to answer questions, further preventing information loss or misinterpretation.

\textbf{Query-oriented Graph Initialization}: 
The document $D$ is divided into $\ermK$ segmentation \{$\rmS_1$, $\rmS_2$, ..., $\rmS_{\ermK}$\}, which is used as the static memory to keep the original information. 
And the structured memory is constructed by the graph $G_k = \{E_k, R_k\}$, where $\rmE_k$ is the set of nodes and $\rmR_k$ is the set of edges.
The entities in the original document, which serve as nodes in the graph, are crucial for identifying the relevant segmentation.
We first prompt LLM to extract the key entities according to question $Q$, then combine with the schema-based tool\footnote{https://stanfordnlp.github.io/CoreNLP} to get the topic entities $\rmE_k = \{e^{'}_{k,1}, e^{'}_{k,2}, ..., e^{'}_{k, N_k}\} $ in $\rmS_k$, where $N_k$ is the number of entities in $\rmS_k$.

The requirement for a fixed schema in Relation Extraction~(RE) significantly constrains domain specificity, making it challenging to adapt to complex scenarios. 
Consequently, we adopt an approach akin to Open Information Extraction~\cite{hybridREJiaSDCX22} for relation extraction. Instead of relying on segments from the original document, descriptive relation representations \{${r}_{k,(i,j)} \in \rmR_k $\} between ${e}_{k, i}$ and ${e}_{k,j}$ are generated through LLM. Thus, the edges in the graph $G_k$ are obtained.

Doubt serves as a fundamental catalyst for the acquisition of knowledge. We advance the methodology of knowledge graph updates through a mechanism predicated on question generation. This mechanism is formalized as $\sQ_k = \text{LLM}(T, \rmS_k, \rmE_k, \rmR_k)$, where $\sQ_k$ denotes the generated question leveraging a Large Language Model (LLM) and the $T$ represents the summary of the entire document.
To ensure the diversity of the generated questions, a novel question is incorporated into the question pool only if its ROUGE-L similarity score with any existing question is less than 0.6.\footnote{Excessive number of questions could introduce further noise, we empirically selected 0.6 as threshold.} Entities and relations are further supplemented through similar techniques based on the questions $\sQ_k$ during the initial stage.

\textbf{Global Graph Combination}: 
In the process of integrating sub-graphs $\{G_1, G_2, ..., G_{\ermK}\}$ into a comprehensive global graph $G$, we further apply the entity disambiguation and relation merging. 
After entity disambiguation, entities associated with distinct segments are further elaborated into triples of the form ${e}_{k,j} = \{{e}^{'}_{k,j}, \sM_{k,j}, \sN_{k,j}\}$, thereby facilitating the update of paragraph indices. 
Here, $\sM_{k,j}$ and $\sN$ represent the sets of entity mentions and segment indices, respectively.
After merging entity pairs ($e_{n,i}, e_{k,i'}$) and ($e_{n,j}, e_{k,j'}$) into consolidated entities $e_{(n,k),i}$ and $e_{(n,k),j}$, 
Following the initial setup phase, we generate questions for merging relationships $Q_s$ according to the entities and their original segments, which could benefit from rethinking the relation between entity pairs.
Consequently, relations $r_{n,(i, j)}$ and $r_{k,(i', j')}$ are amalgamated into a unified relation $r_{(k,n), (i, j)} = \text{LLM}(r_{n,(i, j)}, \rmS_n, r_{k,(i', j')}, \rmS_k, Q_s, Q, T)$. Static segments will be associated with the entity nodes, thereby organizing the connection between the structured and long-term parts in the memory pool, as shown in Figure \ref{fig:ms}.
\vspace{-5pt}

\subsection{Reflection-based Segments Navigation}
To accurately determine the segments relevant to question $Q$, three segment-searching algorithms are proposed to explore the impact of different focuses. The approaches begins by extracting the initial set of entities $\mathbb{E}_{\text{s}}$ and the associated segments $\mathbb{R}_{\text{s}}$ from the query. 
Subsequently, segments corresponding to the initial entity set $\mathbb{E}_{\text{s}}$ are extracted and designed as important segments $\sS_{\text{imp}}$. 

\textbf{\TRIAL{}}.
To enable multiple explorations in complex scenarios and allow the model to find more relevant segments based on previous iterations, we employ a straightforward \TRIAL{} approach.
This method iteratively updates the entity set without the guidance of relations to enhance the relevance of text segments, using a bipartite graph structure composed solely of entities and segment nodes in the memory pool.
After obtaining the initial answer from the model, we utilize its reasoning capability to verify whether the current entity set $\mathbb{E}$ can sufficiently answer the question. If the model cannot answer the question, it iterates by updating $\mathbb{E} \leftarrow LLM(\mathbb{E}, R, \sS_{\text{imp}}, Q)$. This process continues until the model can answer the question or reaches the current window length limit.

\RETRIEVAL{}.
As the number of nodes increases, the model’s ability to filter nodes and the window length both fall short of meeting the need of \TRIAL{}. \RETRIEVAL{} further leverages the structured part of the memory pool, $G$, to enhance exploration from segments.
$\mathbb{E}_s$ is first expanded by incorporating adjacent nodes within the graph. The expansion is guided by the computation of similarity scores between the given query $Q$ and the $\mathbb{R}_s$, retaining nodes $\mathbb{E}'$ with high similarity.
The entity set $\mathbb{E}$ is updated through $\mathbb{E} \gets \mathbb{E}' \cup \mathbb{E}$ following each expansion iteration.
The expansion process continues until either a predefined threshold of expansion iterations is reached or no new edges with sufficiently high similarity scores are found. 
The resultant expanded set of entities $\mathbb{E}$ is then to generate the elaborated query through $\sQ = LLM(\mathbb{E}, \mathbb{R}, Q)$. 
$\sQ$ is utilized to retrieve relevant segments, which are constructed into $c$ to subsequently address $Q$. $c$ is the concatenation of the original question and the generated question used to retrieve all relations.

\textbf{\ours{}}.
To trace the experience across multiple iterations, \ours{} utilizes a reflection-based navigation strategy that dynamically traverses the graph $G$, which is shown in Figure \ref{fig:navi}.
Initially, \ours{} sets up an additional segmentation set, $\sS_{\text{add}}$, as $\emptyset$ for further navigation.
For each iteration, the real relevant segments $\sS_{\text{mix}}$ is formed by combination of $\sS_{\text{imp}}$ and $\sS_{\text{add}}$. Then the LLM is employed to evaluate whether $\sS_{\text{mix}}$ can answer the question $Q$ by the feedback of the question $Q$ and the reason $R$ for the failure is identified if not.
While the $\sS_{\text{mix}}$ is insufficient for the current query, the first-order adjacent nodes $\mathbb{E}_{\text{adj}}$ of the current entities and all connecting edges $\mathbb{R}$ are collected.
Drawing upon the lessons learned from the previous iteration, the query, reason, and current entities $[Q; R;\mathbb{E}_{\text{s}}]$ are used to select the next expanded node.
Contriever\cite{Izacard22contriever} is used for calculating the similarity score between the $[Q; R;\mathbb{E}_{\text{s}}]$ and each edge in $\mathbb{R}$.
After the most relevant adjust node $e$ is selected and merged with the new entity $e$ into the entity set $\mathbb{E}$, the segments corresponding to the new entity $e$ are collected. 
All segments not present in $\sS_{\text{mix}}$ are added to $\sS_{\text{add}}$ with the similarity score.
Allowing for thorough updates to \(\sS_{\text{add}}\), a crucial step is to ensure that the context window remains sufficient after updates. 
\(\sS_{\text{add}}\) is processed and filtered according to the similarity score while adhering to the window constraints.
This iterative segment combination, window checking, question evaluation, and sub-graph expansion process continues until the question $Q$ is satisfactorily answered.
The detail of the algorithm is shown in Algorithm \ref{alg:navi}.
\section{Experiment}
\subsection{Baselines}
To evaluate the effectiveness of the proposed method, we compare the \ours{} with several methods. 
\textbf{Retrieval Methods}. We use dense retrievers Contriever~\cite{Izacard22contriever} and BM25 for all datasets.
\textbf{Vanilla}, we use the model with 4096 context windows, including LLaMa\-70B\-Chat~\cite{llama2} and GPT-3.5-turbo. 
\textbf{Summary}, we summarize the long context to fit the context window size.
\textbf{DECOM}\cite{fu-etal-2021-decomposing-complex} simplifies complex questions into sub-questions, answers them sequentially.
\textbf{LLMLingua}~\cite{pan2024llmlingua2}, compresses prompts to speed up inference and enhance response efficiency in large language models.
\textbf{Long context model}, including Vicuna-7b-32k, Vicuna-13b-16k, Longchat-7b-32k, Longchat-13b-16k,Llama-70b-32k.
\textbf{\textsc{MemWalker}}~\cite{chen2023walking}, which utilize a tree of summary nodes.

\subsection{Experiment Setting}

The official LLaMa-70B-Chat and GPT-3.5-turbo are used for the majority of our experiments.  The two models both have a maximum 4,096 token context length. We utilize the model in a zero-shot prompting fashion without further fine-tuning or using few-shot examples for our tasks. For both memory tree construction and generating action and reasoning for navigation, we use top-p sampling. The temperature is initially set to 0, but if the output cannot be parsed, we increase the temperature to 0.7 and retry up to four times. The segment size is set to 600 for all datasets. 
\begin{table*}[t]
\centering
\footnotesize
\setlength{\tabcolsep}{3.3mm}{
\begin{tabular}{l|c|cc|cc|cc}
\toprule
\multicolumn{1}{c}{}  & \multicolumn{1}{c}{QuALITY} & \multicolumn{2}{c}{HotpotQA} & \multicolumn{2}{c}{2WikiMultihopQA}   & \multicolumn{2}{c}{MuSiQue}   \\ \cmidrule(lr){2-2} \cmidrule(lr){3-4} \cmidrule(lr){5-6} \cmidrule(lr){7-8} 
\multicolumn{1}{c}{\multirow{-2}{*}{Method}} & \multicolumn{1}{c}{Acc}  & \multicolumn{1}{c}{EM} & \multicolumn{1}{c}{F1} & \multicolumn{1}{c}{EM} & \multicolumn{1}{c}{F1} & \multicolumn{1}{c}{EM} & \multicolumn{1}{c}{F1} \\ \midrule
\multicolumn{8}{c}{\cellcolor[HTML]{E0DEDE}Long Context Model}   \\
Vicuna-7b-16k & 0.4175 & 0.2750 & 0.2316 & 0.2750 & 0.1779 & 0.0700 & 0.0928 \\
Vicuna-13b-16k   & 0.5191 & 0.2800  & 0.3074 & 0.3400 & 0.2674  & 0.1150   & 0.1315 \\
Longchat-7b-32k &0.2938  &0.3750 &0.2932 &0.3750 &0.2012 &0.1400 &0.1256 \\
Longchat-13b-16k   & 0.3250 & 0.3000  & 0.2465   & 0.2850   & 0.1945 & 0.1100 & 0.1335 \\
Llama-2-70b-32k   & \underline{0.5765} & \textbf{0.4000}  & \underline{0.3372} & \textbf{0.4200} & \underline{0.3725} & \textbf{0.1750}   & \underline{0.1605}  \\ 
\midrule
GPT-3.5-16k   &\textbf{0.7162}  & \underline{0.3900} & \textbf{0.4415 }& \underline{0.4050} & \textbf{0.4278} &\underline{0.1550} & \textbf{0.1888} \\
\midrule
\multicolumn{8}{c}{\cellcolor[HTML]{E0DEDE}LLaMa-70B-Chat } \\
BM25 (top 3)  & 0.6055 & \textbf{0.4800} & \textbf{0.4418}  & 0.4000  & 0.3285  & 0.1750   & 0.1697  \\
Contriever (top 3) &0.6098   & 0.3900 & 0.3552  & 0.4500 & \textbf{0.3632}  & 0.1300 & 0.1640   \\
Vanilla~(keep left)   & 0.6131 &  0.3750 & 0.2966 & 0.4050 & 0.2393& 0.1000 & 0.0898 \\
Vanilla~(keep right)  & 0.6251 &0.2950 &0.2655 & 0.3400& 0.2185& 0.1200& 0.1202\\
\textsc{MemWalker}\dag   & 0.6302 &0.4350 &0.2046 &0.3850 &\underline{0.3462}  &0.1600 &0.1628 \\
\textsc{LLMLingua} &0.4636 & 0.2750 & 0.2662&  0.2750 & 0.2439& 0.0600& 0.0334\\
\textsc{DECOM} &0.4515&0.4250& 0.3938& 0.3350& 0.3136 &0.1300 &0.1277\\
\ours{} (TE) &0.6351    & - & - & - & - & - &- \\ 
\ours{} (GES) &\underline{0.6441}  &0.4550 &0.4032 &\underline{0.4650} &0.3044 &\underline{0.1800} & \underline{0.1933} \\ 
\ours{} (KN)   &\textbf{0.6531}   &\underline{0.4650} & \underline{0.4211} &\textbf{0.4700} &0.3314 &\textbf{0.2000} &\textbf{0.1996} \\ 
\midrule
\multicolumn{8}{c}{\cellcolor[HTML]{E0DEDE}GPT-3.5-turbo} \\
BM25 (top 3) & 0.6359 & 0.3850   & 0.4511  & 0.3200 & 0.3471  & 0.0850   & 0.1476  \\
Contriever (top 3) & 0.6275  & 0.3650   & 0.3950  & 0.4100 & \underline{0.4523}  & 0.1350   & \underline{0.1948}  \\
Vanilla (keep left)  & 0.6448 &0.3650 &0.4229 &0.3350 &0.3615 &0.0650 &0.1020 \\
Vanilla (keep right) & 0.6687 &0.2550 &0.3098 &0.2550 &0.2691 &0.0750 &0.1213 \\
\textsc{MemWalker}\dag   & 0.6622  &0.3350 & 0.4113 &0.4050 &0.4432 &0.0950 &0.1323 \\
\textsc{LLMLingua} &0.5360 & 0.3200 & 0.3393&  0.2600 & 0.2817& 0.1000& 0.0811\\
\textsc{DECOM} &0.5495 & 0.3850 & 0.3933& 0.3750& 0.3939 &0.1000 &0.1086\\
\ours{} (TE)  &\underline{0.6936}  &- &- &- &- &- &- \\ 
\ours{} (GES)  &0.6801   &0\underline{.4150} & \textbf{0.4731} &\underline{0.4250} &0.4432 & \underline{0.1850} & 0.1931 \\ 
\ours{}(KN)   &\textbf{0.7207}   &\textbf{0.4200} &\underline{0.4632} &\textbf{0.4250} &\textbf{0.4698} & \textbf{0.1900} & \textbf{0.2094}\\ 
\bottomrule
\end{tabular}
}
\caption{Mian results on QuALITY and Multi-doc QA datasets, \dag denotes that the results are reproduced based on the original paper. TE denotes the \TRIAL{}, GES denotes the \RETRIEVAL{}, and KN denotes the \ours{}. We bold the highest scores and underline the second-highest scores for each block.}
\vspace{-14pt}
\label{tab:main_result}
\end{table*}

\subsection{Dataset \& Metrics}
The evaluation datasets predominantly comprise sections on multiple-choice questions (MCQ) and multi-hop question-answering (Multi-Doc QA). 
\textbf{QuALITY}~\cite{pang-etal-2022-quality} is employed for the MCQ component,
which comprises extended narratives derived from Project Gutenberg, coupled with questions annotated by expert human annotators. For our experimental analysis, we selected a subset consisting of 222 examples, and accuracy is used for evaluation.
To evaluate the capability of short-context models in comprehending extended texts, we opted against using the original multi-hop QA dataset. Instead, we employed \textbf{Multi-Doc QA} derived from LongBench~\cite{bai2023longbench}, with an average length exceeding 8,000 tokens, ensuring suitability for evaluating the understanding of extensive textual information.
The datasets incorporate the Wikipedia passages encompassed version of HotpotQA~\cite{yang-etal-2018-hotpotqa}, 2WikiMultihopQA~\cite{ho-etal-2020-constructing} and MuSiQue~\cite{trivedi-etal-2022-musique}. Consistent with previous works~\cite{yang-etal-2018-hotpotqa, bai2023longbench}, exact match~(EM) and $\ermF1$ is selected for evaluation.

\subsection{Main Result}

We evaluate the proposed model \ours{} under the long context question-answer setting. Table \ref{tab:main_result} shows the comparison results of \ours{} with different navigation methods on QuALITY and three Multi-doc QA datasets against all baseline methods. \TRIAL{} is only tested on QuALITY due to the excessive number of entities present on Multi-Doc QA.
We can see that:

1) Our study presents significant advancements over retrieval methods on the QuALITY dataset, on both the LLaMa-70B-Chat and GPT-3.5-turbo models. Detailed results related to retrieval methods are referred to in Appendix \ref{app:retrieve}. 
Notably, our approach, when compared to models utilizing longer context windows, outperforms other models designed for long-text comprehension, including GPT-3.5-16k.
This highlights the effectiveness of our method in selecting relevant segments,
promoting a comprehensive understanding of complex documents.

2) Furthermore, our methods demonstrated exceptional performance on Multi-Doc QA datasets. 
Specifically, on the 2WikiMultihopQA and MuSiQue datasets, our approach achieved remarkable results and yields comparable results to existing retrieval models on the HotpotQA dataset. Even though BM25 obtained a satisfactory result on HotpotQA, it still struggles with complex tasks like QuALITY and MuSiQue.
Across all three datasets, our method outperformed long-document comprehension models. This indicates our approach's adaptability to multi-hop tasks that require the understanding of entity relationships, capturing these relationships more effectively than long-document models.
The lesser improvement on fewer inference steps HotpotQA questions and more significant advancements on the complex MuSiQue and QuALITY tasks illustrate our method's effectiveness in organizing original paragraphs for complex tasks.
GPT-3.5-16k tends to focus more on the overall context, while GPT-3.5-turbo pays closer attention to details. This is also why GPT-3.5-16k shows better performance on QuALITY datasets.
\vspace{-1pt}

3) Three strategies proposed in our study outperformed retrieval-enhanced and direct truncation strategies. 
\TRIAL{}'s performance improvement demonstrates the effectiveness of iterative exploration. By using entities as anchors and leveraging the reflective iterative exploration of segments, it shows the impact of reflective iteration in acquiring relevant segments. 
\RETRIEVAL{}, although lacking an iterative navigation process, still utilizes structured memory to guide the LLM. This indicates that our query-based structured memory provides better guidance and relevance compared to predefined indexing. 
\ours{} achieves the best performance. It combines the strengths of both \TRIAL{} and \RETRIEVAL{}, using the graph to guide the model and leveraging past errors to inform future steps. This approach allows the model to utilize specific structured knowledge while retaining the ability to iteratively combine different segments.
\vspace{-6pt}

\subsection{Ablation Study}
\begin{table}[htbp]
\centering
\resizebox{0.46\textwidth}{!}
{
\begin{tabular}{lccc}
\toprule
Method             & Overall & Easy      & Difficult \\
\midrule
\ours{} & 0.7207  & 0.7000    & 0.7450 \\
\quad -w/o \textsc{Graph Update}     & 0.6846  & 0.7166    & 0.6470 \\
\quad -w/o \textsc{Open Entity}     & 0.6621  & 0.7083    & 0.6078 \\
\quad -w/o \textsc{Reflection}     & 0.6801  & 0.6833    & 0.6764 \\
\quad -w/o \textsc{Navigation}      & 0.6351  & 0.6750    & 0.5882 \\
\bottomrule
\end{tabular}}
\vspace{-4pt}
\caption{Ablation study on Quality dataset through GPT-3.5-turbo. Quality is annotated with difficult levels, and we report the results on different levels.}
\label{tab:ab}
\vspace{-8pt}
\end{table}

\setlength{\tabcolsep}{2.3mm}
In this section, we study the effectiveness of each proposed module by removing them from \ours{} on QuALITY Datasets. 
Results in Table \ref{tab:ab} show that removing any component decreases performance, which observe their individual contributions to performance.
\textsc{Graph Update} denotes the additional question generation and graph update during initialization. \textsc{Open Entity} refers to the entity recognition method through LLM. When these two components are removed, we observed improved performance on the easy level but a significant decline in performance on the difficult level. This phenomenon indicates that query-oriented structured memory is crucial for addressing complex problems.
\textsc{Reflection} involves adding reasoning as a reflection when calculating similarity with relations. Without leveraging past experiences and only relying on the query to traverse the graph, there is a significant loss in performance. This demonstrates that our reflection mechanism provides more precise guidance.
\textsc{Navigation} refers to the navigation stage. Without \textsc{Navigation}, we simply use the start entity to locate the segments and then answer the query. This approach loses the iterative navigation process, resulting in an overall decline in performance. Our technique of reorganizing paragraphs helps to mitigate this issue and compensates for this deficiency.
\vspace{-4pt}

\subsection{Performance over Different Max Trials}
\vspace{-4pt}
\begin{figure}[h]
    \centering
    \includegraphics[width=0.4\textwidth]{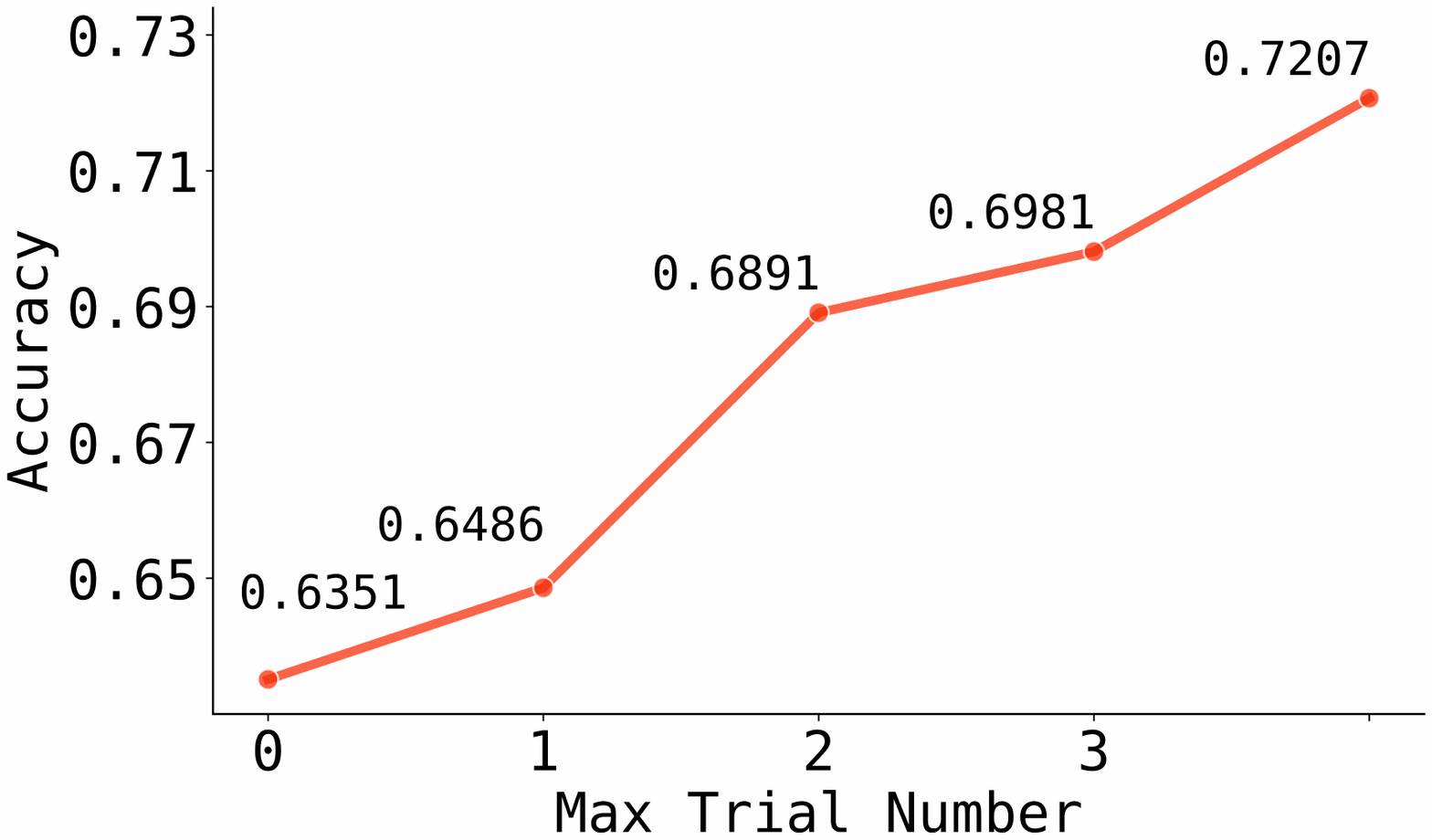}
    \vspace{-10pt}
    \caption{Accuracy on QuALITY datasets with different max trials number.}
    \label{fig:max}
    \vspace{-10pt}
\end{figure}
In this study, we introduce a constraint on the number of updates allowed for entity set $\mathbb{E}$ during the search process, which is shown as Figure~\ref{fig:max}.
Once the maximum number of updates is reached, the search is terminated, then the answer is generated based on the current state of $\mathbb{E}$.
Our results indicate a significant enhancement in performance transitioning from the first to the second update, underscoring the effectiveness of the navigation strategy.
Upon undergoing a series of iterative trial and error processes, it has been observed that the performance of the model reaches a plateau, indicating a deceleration in the rate of improvement. The model could still gain profit from the navigation. Experimental observations indicate that trail number could be a trade-off between efficiency and effectiveness.
\vspace{-3pt}

\subsection{The impact of irrelevant information}
\begin{table}[htbp]
\centering
\resizebox{0.47\textwidth}{!}{
\begin{tabular}{l|cc|cc|cc}
\toprule
\multicolumn{1}{c}{}   & \multicolumn{2}{c}{Head} & \multicolumn{2}{c}{Middle}   & \multicolumn{2}{c}{Tail}   \\ 
\cmidrule(lr){2-3} \cmidrule(lr){4-5} \cmidrule(lr){6-7}
\multicolumn{1}{c}{\multirow{-2}{*}{Datasets}} &  \multicolumn{1}{c}{EM} & \multicolumn{1}{c}{F1} & \multicolumn{1}{c}{EM} & \multicolumn{1}{c}{F1} & \multicolumn{1}{c}{EM} & \multicolumn{1}{c}{F1} \\ 
\midrule
\multicolumn{7}{c}{\cellcolor[HTML]{E0DEDE}Vicuna-7b-16k} \\
Hotpot  & 0.3150&	0.2283&  0.2650&	0.1889& 0.3250&	0.2257\\
2Wiki   & 0.3850&	0.2387& 0.2700&	0.1749& 0.3150	&0.2043\\
Musique & 0.0750 & 0.0896 & 0.0500 & 0.0733&  0.1950 & 	0.2060\\
\midrule
\multicolumn{7}{c}{\cellcolor[HTML]{E0DEDE}Vicuna-13b-16k} \\
Hotpot  &0.3750	& 0.3444 &0.3050	& 0.2643 &0.3650	& 0.3282\\
2Wiki  & 0.3400 &	0.2584& 0.3550 &	0.2864& 0.3900	& 0.3052\\
Musique& 0.0900& 0.0783&  0.0700& 	0.0680& 0.1000	&0.0725\\
\bottomrule
\end{tabular}}
\vspace{-4pt}
\caption{Result on Multi-Doc QA datasets, the supporting segments are positioned at the beginning, middle, and tail of the document.}
\label{tab:ps}
\vspace{-10pt}
\end{table}

To further investigate the influence of irrelevant documents, we conducted additional tests on cross-document multi-hop questions. In this section, we retained the original wiki support text from the Long Bench dataset without chunking the long text, enabling us to more clearly identify the impact of irrelevant documents.
Following \cite{liu2023lost}, we investigated the impact of varying the placement of supporting segments within a document, and the result is shown in Table \ref{tab:ps}.
Specifically, we positioned the corresponding support segments at the beginning, middle, and tail of the overall document. 
Our findings are consistent with those reported by \citet{liu2023lost}, indicating that the placement of information can significantly affect the model's performance on the Multi-Doc QA task.
Therefore, in our subsequent experiments, we placed the multi-hop support text at both the beginning and end of the document, ensuring that the supporting segments remain in these positions to avoid any positional impact.

\begin{figure}[h]
    \centering
    \subfigure[HotpotQA]{
    \includegraphics[width=0.22\textwidth]{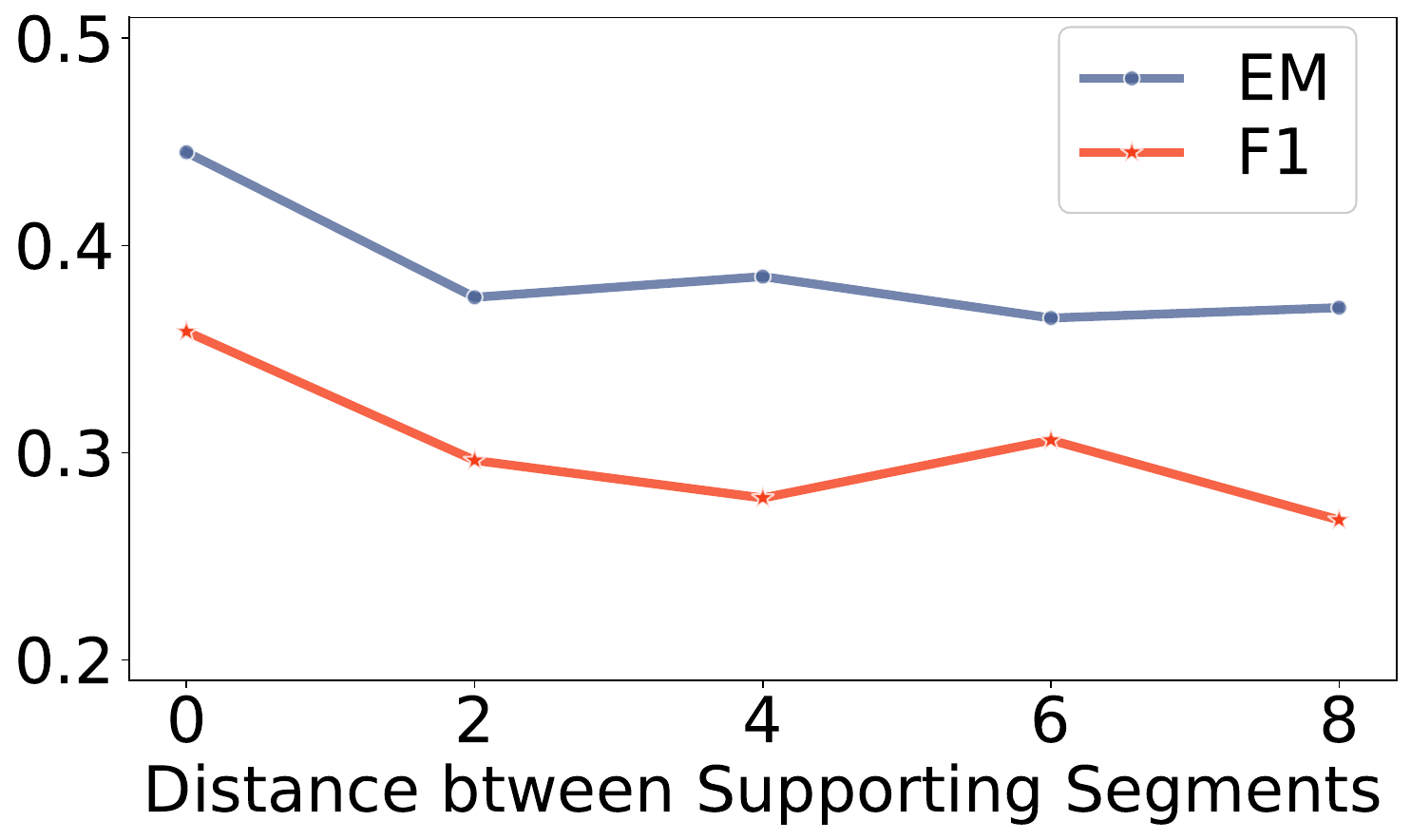}
    }
    \subfigure[MuSiQue]{
    \includegraphics[width=0.22\textwidth]{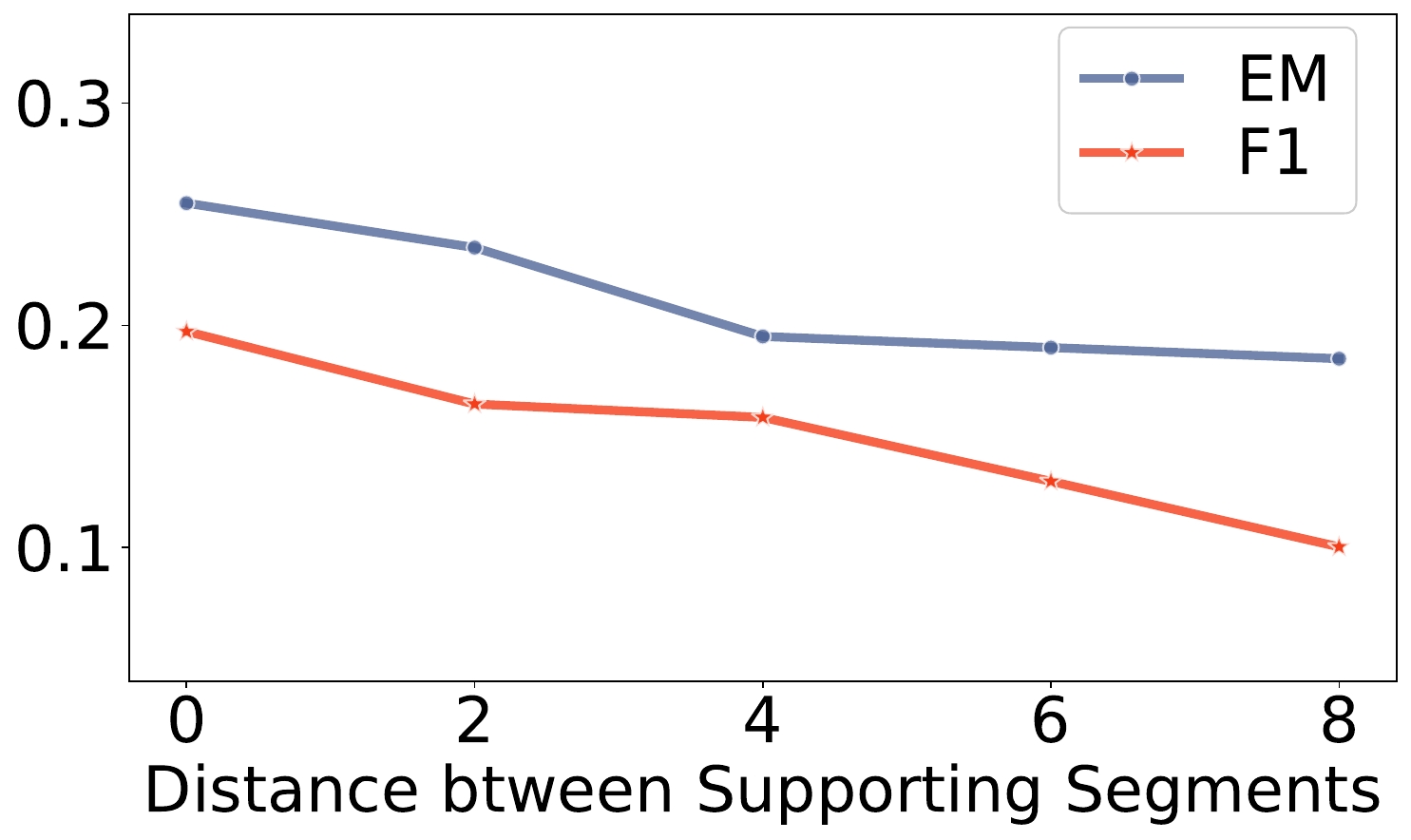}
    }
    \vspace{-10pt}
    \caption{The performance over different distances between the supporting segments.}
    \label{fig:set2}
\vspace{-10pt}
\end{figure}
Irrelevant documents utilized to increase the distance are retained, represented as [$S_1; S_{non}; S_2$], where $S_1$ and $S_2$ are the supporting segments, and $S_{non}$ represent sequences of non-relevant segments.
The outcomes are depicted in Figure~\ref{fig:set2}.
The results indicate that although the model exhibits heightened attention to the positions at the beginning and end of the text, its performance is still impacted by the distance between documents. 
Effectiveness consistently declines as distance increases on both the HotpotQA and MuSiQue datasets.

In summary, tasks of increasing complexity necessitate the aggregation of relevant documents, where the effects can even surpass the influence of the locations of the supporting documents.
Long context understanding requires high capability in capturing long-distance dependencies that demand complex reasoning. The case study in Appendix~\ref{app:case} illustrates our ability to find relevant documents in complex situations.
\ours{} constructs structured knowledge for each segment, mitigating the impact of long-distance dependencies.
By utilizing structured knowledge to reorganize the segments, our approach facilitates the close integration of text with dependency relationships,
which leads to superior effectiveness on Multi-Doc tasks compared to the long context windows model.

\section{Related Work}
\subsection{Memory Mechanism}
Except for the LLMs with extended context window size \citep{PressSL22, guo2022longt5, ainslie2023colt5} and vanilla retrieve augment generation methods \cite{shi2023replug, zhang2023merging, yu2023chain, chevalier2023adapting},
numerous attempts have been made to enhance the memory capabilities of neural models\cite{chen2023walking}.
Memory mechanisms could be predefined representations, including raw natural languages \cite{mmbankZhongGGYW24, chatdb2023hu}, structured information \cite{chen2023walking} such as tuples, databases, etc., and paramterized representations \cite{genread2023yu,ab2024tang}.
To leverage the constructed memory, the method stores all the information of the agent-environment interaction history based on long-context strategies, maintains the most recently acquired memories \cite{scm2023liang,mengpt2023Packer,ebbinghaus2013memory}, and selects memory contents based on their relevance, importance, and topics \cite{mmbankZhongGGYW24,gaParkOCMLB23}. 
However, these memory-based algorithms do not allow the preparation of specific memory structures for different queries, and further memory integration is also ignored. We use the Knowledge-directed Navigation method to reorganize the context adjustably.
\vspace{-8pt}

\subsection{KG augmented Language Modeling}

Knowledge Graphs (KGs) offer dynamic, structured knowledge representation, enhancing Large Language Models (LLMs) through explicit structured knowledge, as demonstrated by early studies such as \citep{zhang2019ernie,Knowbert,LUKE,wang2021kepler,K-Adapter, KELLMsurvey}. 
Methods include generating SPARQL query backbones \citep{DB-BLINDER}, sampling entity-related triples \citep{knowledgeaugmented}, decomposing \citep{cok}, and employing a retriever-reader-verifier\citep{wang2023knowledgedriven}. Other approaches like \citep{structgpt} enable LLMs to navigate KGs through greedy search, indicating a trend towards interactive KG exploration and reasoning \citep{sun2023thinkongraph}, graph-driven context retrieval \citep{banerjee2024context}, and the formulation of examinations to assess LLM comprehension \citep{wan2024mitigating}.
Previous methods primarily incorporate graphs into the response in various ways or introduce external knowledge graphs. Our approach directly utilizes the original documents to construct graphs and employs the original paragraph information during the answer generation process. This strategy effectively circumvents the propagation of errors that may arise during the extraction process.

\vspace{-3pt}
\section{Conclusion}
\vspace{-3pt}

In this study, we propose the Question-then-Reflection Memory mechanism, which enables the utilization of structured knowledge to guide the mining of supporting segments. 
\ours{} aligns with the query’s needs by using structured memory for guidance, and original segments are included as long-term static memory to avoid information bias.
We introduce \TRIAL{}, \RETRIEVAL{}, and \ours{} as methodologies to navigate through structured memory and find relevant segments. 
\ours{} reflects on errors from past experiences during multiple iterations, allowing the model to learn from its mistakes and improve performance. 
Experimental results on MCQ and Multi-doc QA demonstrate that our model achieves superior results with various methods, validating the effectiveness of our approach. 
\section{Limitations}
In this paper, we propose a Question then Reflection Memory Mechanism method for iteratively select the relevant segments.
The limitations of the proposed method are as follows:
1) The computational cost of initializing and updating the knowledge in processing the long context, may restrict its system applicability in scenarios where there is a need to quickly build additional knowledge bases.
2) Given the constraints imposed by the size of the model window, reliance on entity resolution is utilized to bridge distinct paragraphs during the extraction of pertinent knowledge in constructing the structure knowledge. The segmentation of text into paragraphs potentially impacts the efficacy of connections within the graph's structure. 

In future work, we plan to mitigate the method's computational cost by developing more efficient pipelines. 
We aim to limit the updates of the graph to a very small local range to reduce the cost of updating the graph structure or explore the incorporation of external knowledge sources to mitigate this limitation and enhance the model's performance.

\section{Ethics Statement}
This work was conducted in rigorous compliance with the ACL Ethics Policy. All datasets and large language models (LLMs) used for evaluation are publicly available. 
Furthermore, our work strives to explore a Question and Reflection Memory Me for long-context understanding. 
We do not foresee any form of negative ethical impact induced by our work.

\section{Acknowledgement}
We sincerely thank the anonymous reviewers for their helpful feedback and the conference committee for their hard work. This work was supported by the Joint Fund of the National Natural Science Foundation of China (Grant No. U21B2009). 
\bibliography{custom}
\clearpage
\appendix
\appendix
\onecolumn
\section{Navigation strategy}
\label{app:navi}
To better describe our navigation strategy, the detailed algorithm of navigation is shown as Algorithm \ref{alg:navi}
\begin{algorithm}[htbp]
\caption{Reflection-Based Navigation Strategy}
\label{alg:navi}
\begin{algorithmic}[1]
\REQUIRE Query $Q$, Initial Entity Set $\mathbb{E}_{\text{s}}$, Initial Relation Set $\mathbb{R}_{\text{s}}$, Segment Set $\sS_{\text{add}} \leftarrow \emptyset$, Context Windows Size $L$
\ENSURE Answer to the question $Q$
\STATE $\sS_{\text{imp}} \leftarrow$ Extract segments corresponding to $\mathbb{E}_{\text{s}}$
\STATE $\sS_{\text{mix}} \leftarrow \sS_{\text{imp}} \cup \sS_{\text{add}}$
\WHILE{not $\text{answered}(Q)$}
    \IF{LLM can answer $Q$ with $\sS_{\text{mix}}$}
        \RETURN Answer
    \ELSE
        \STATE $R \leftarrow \text{LLM}(\sS_{\text{mix}}, Q) $
        \STATE $\mathbb{E}_{\text{adj}} \leftarrow$ First-order adjacent nodes of $\mathbb{E}_{\text{s}}$
        \STATE $\mathbb{R} \leftarrow$ Connecting edges of $\mathbb{E}_{\text{adj}}$
        \STATE $S \leftarrow \text{Similarity}([Q; R; \mathbb{E}_{\text{s}}], \mathbb{R})$
        \STATE $e \leftarrow \text{Max}_S(\mathbb{R})$
        \STATE $\mathbb{E} \leftarrow \mathbb{E} \cup \{e\}$
        \STATE $\sS_{\text{new}} \leftarrow$ Extract segments corresponding to $e$
        \STATE $\sS_{\text{add}} \leftarrow \sS_{\text{add}} \cup (\sS_{\text{new}} \setminus \sS_{\text{mix}})$
    \ENDIF
    \STATE $\sS_{\text{mix}} \leftarrow \sS_{\text{imp}} \cup \sS_{\text{add}}$
    \IF {LEN($\sS_{\text{mix}}$) $>$ $L$}
        \STATE  $\sS_{\text{add}} \leftarrow$ Filter $\sS_{\text{add}}$ according to $S$
        \STATE $\sS_{\text{mix}} \leftarrow \sS_{\text{imp}} \cup \sS_{\text{add}}$
    \ENDIF
\ENDWHILE
\end{algorithmic}
\end{algorithm}

\section{Retrieval Result}
\label{app:retrieve}
We progressively incorporated the content retrieved into the original document in the retrieval experiments. "Max" indicates including retrieved content in the prompt until the length limit is reached. For the multi-doc QA dataset, we reported Exact Match (EM), and for the QuALITY dataset, we reported Accuracy. The results are shown in the table \ref{tab:retri}. Since performance does not improve with the inclusion of more data, we selected the top 3  with the higher average values for comparison in our paper.

\begin{table}[htbp]
\centering
\caption{Results about the Retrieval Methods}
\label{tab:retri}
\begin{tabular}{lllll}
\toprule
{\color[HTML]{333333} \textbf{Method}}   & \multicolumn{1}{c}{{\color[HTML]{333333} \textbf{Hotpot}}} & \multicolumn{1}{c}{{\color[HTML]{333333} \textbf{2wiki}}} & \multicolumn{1}{c}{{\color[HTML]{333333} \textbf{Musique}}} & \multicolumn{1}{c}{{\color[HTML]{333333} \textbf{Quality}}} \\
\midrule
{\color[HTML]{333333} BM25(top 1)}       & {\color[HTML]{333333} 0.3800}                                & {\color[HTML]{333333} 0.2450}                              & {\color[HTML]{333333} 0.1150}                                & {\color[HTML]{333333} 0.5288}                               \\
\rowcolor[HTML]{F8F8F8} 
{\color[HTML]{333333} BM25(top 2)}       & {\color[HTML]{333333} 0.4450}                               & {\color[HTML]{333333} 0.3600}                               & {\color[HTML]{333333} 0.1500}                                 & {\color[HTML]{333333} 0.5705}                               \\
{\color[HTML]{333333} BM25(top 3)}       & {\color[HTML]{333333} 0.4800}                                & {\color[HTML]{333333} 0.4000}                                & {\color[HTML]{333333} 0.1750}                                & {\color[HTML]{333333} 0.6055}                               \\
\rowcolor[HTML]{F8F8F8} 
{\color[HTML]{333333} BM25(Max)}         & {\color[HTML]{333333} 0.4550}                               & {\color[HTML]{333333} 0.4050}                              & {\color[HTML]{333333} 0.1750}                                & {\color[HTML]{333333} 0.6208}                               \\
\midrule
{\color[HTML]{333333} Contriever(top 1)} & {\color[HTML]{333333} 0.3350}                               & {\color[HTML]{333333} 0.2800}                               & {\color[HTML]{333333} 0.1000}                                  & {\color[HTML]{333333} 0.5292}                               \\
\rowcolor[HTML]{F8F8F8} 
{\color[HTML]{333333} Contriever(top 1)} & {\color[HTML]{333333} 0.3850}                               & {\color[HTML]{333333} 0.3950}                              & {\color[HTML]{333333} 0.1600}                                 & {\color[HTML]{333333} 0.5959}                               \\
{\color[HTML]{333333} Contriever(top 3)} & {\color[HTML]{333333} 0.3900}                                & {\color[HTML]{333333} 0.4600}                               & {\color[HTML]{333333} 0.1300}                                 & {\color[HTML]{333333} 0.6098}                               \\
\rowcolor[HTML]{F8F8F8} 
{\color[HTML]{333333} Contriever(max)}   & {\color[HTML]{333333} 0.4300}                                & {\color[HTML]{333333} 0.4450}                              & {\color[HTML]{333333} 0.1550}                                & {\color[HTML]{333333} 0.6385}                               \\
\midrule
{\color[HTML]{333333} RKDN(ours)}        & {\color[HTML]{333333} 0.4650}                               & {\color[HTML]{333333} 0.4700}                               & {\color[HTML]{333333} 0.2000}                                  & {\color[HTML]{333333} 0.6531}\\
\bottomrule
\end{tabular}%
\end{table}

\section{Case Study}
\label{app:case}
As illustrated in Table~\ref{tab:case_study_part}, when crucial entities are absent in the query, we can employ an iterative navigation process to search for and identify the relevant entities gradually. Although irrelevant segments may be included, the noise introduced is less than that of retrieval methods. The first two segments rank lower in retrieval, indicating that for complex reasoning tasks, it is essential to construct memory based on the query and reorganize the text to bridge the gap with the query. This validates the effectiveness of \ours{}.
\begin{table*}[htbp]
\centering \footnotesize
\begin{tabular}{p{40em}|>{\centering\arraybackslash}m{3em}|>{\centering\arraybackslash}m{3em}}
\toprule
Segments & Rank & Contain\\
\midrule
Valencia CF had a successful season, finishing in the top four of La Liga and thus qualifying for the UEFA Champions League, thanks to the extension of the competition to include more teams from the top leagues. Valencia also won the Copa del Rey, ending a long trophy drought and marking a successful end to Italian coach Claudio Ranieri’s first spell at the club. & 7 &  \includegraphics[scale=0.5]{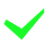}  \\
\midrule
José Daniel Valencia (born 3 October 1955) is an Argentine former professional footballer who played as an attacking midfielder. He is perhaps most famous for having been part of the 1978 World Cup winning squad.Valencia started his club career at Gimnasia y Esgrima de Jujuy but was soon transferred to Talleres de Córdoba, the club at which he would play most of his career.
At Talleres, Valencia suffered the disappointment of finishing runner-up in Nacional 1977, finishing third in Metropolitano 1980, and losing the semi-finals on four occasions. & 5 & \includegraphics[scale=0.5]{fig/correct.png}  \\
\midrule
Higinio Ortúzar Santamaría (10 January 1915 – 8 November 1982) was a Chilean footballer who made his entire career in Spain. The first Chilean in Spanish football, he made his debut for Erandio Club in 1935, and next he played for Barakaldo CF, Athletic Bilbao, Valencia CF, Real Valladolid, and Real Sociedad. He was loaned to Racing de Santander in 1936 for 4,500 pesetas, but he couldn’t play due to the Spanish coup of July. & 1 & \includegraphics[scale=0.5]{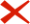}  \\
\midrule
José Raúl Aveiro Lamas (born 18 July 1936) is a Paraguayan former professional footballer who played as a striker. Born in Asunción, Aveiro played for Sportivo Luqueño, Valencia, Valencia Mestalla, Elche, Ontinyent, and Constància. He was also a member of the Paraguay national team between 1957 and 1959. & 2 & \includegraphics[scale=0.5]{fig/wrong.png}  \\
\midrule
Claudio Javier López (Spanish pronunciation:, born 17 July 1974) is an Argentine former footballer, who played as a forward. Nicknamed Piojo (louse), he is best known for his spells with Valencia in Spain and Lazio in Italy. López also had a notable impact in the Argentina national team, participating in two World Cups. & 3 & \includegraphics[scale=0.5]{fig/correct.png}  \\
\midrule
\multicolumn{3}{l}{\textbf{Question}: Which retired Argentine footballer who played as a forward was a main player for Valencia CF?}\\
\multicolumn{3}{l}{\textbf{Golden Answer}: Claudio Javier López}\\
\multicolumn{3}{l}{\textbf{Retriever Answer}: Mario Alberto Kempes Chiodi}\\
\multicolumn{3}{l}{\textbf{Compressed Answer}: Mario Alberto Kempes Chiodi}\\
\multicolumn{3}{l}{\textbf{\ours{} Answer}: Claudio Javier López }\\

\bottomrule
\end{tabular}%
\caption{Case studies of solving Mulri-Doc QA task. For the question, we show the rank of each segment. ``\includegraphics[scale=0.5]{fig/correct.png}" denotes the segment is contained in the final set of \ours{}, ``\includegraphics[scale=0.5]{fig/wrong.png}" denotes the segment is not contained in the final set of \ours{}.}
\label{tab:case_study_part}%
\end{table*}%


\section{Computational Complexity Analysis}

We compared various methods for handling long texts using several requests, using the GPT-3.5 model to evaluate performance on QuALITY metrics. 
The experiments were conducted with a window size of 4096.
\textsc{SlidingWindow} apply a sliding window in five requests to ensure the entire text is covered. 
Our method achieved optimal results in comparison to others, as summarized in Table~\ref{table:comparison}.

\begin{table}[htbp]
\centering
\caption{Computational Complexity Comparison of Different Methods for Long Text Processing}
\begin{tabular}{lcc}
\toprule
\textbf{Method} & \textbf{QuALITY} & \textbf{Number of Requests} \\ 
\midrule
\textsc{GIST}~\cite{lee2024human} & 0.7005 & 4.32 \\ 
\textsc{MemWalker}~\cite{chen2023walking} & 0.6622 & 3.71 \\ 
\textsc{SlidingWindow} & 0.6891 & 5.00 \\ 
\ours{} & 0.7207 & 3.96 \\ 
\bottomrule
\end{tabular}
\label{table:comparison}
\end{table}

\section{Prompt in QRMeM}
\label{app:prompt}
The prompt used in this paper is shown as follows, the \blue{blue} text denotes the part to be filled with the different documents.


\begin{prompt}[title={The prompt used for asking the LLM weather could answer the question}, label=prompt:answer]
Read the text in triple quotes and answer a question:\\
\texttt{"""} \\
\blue{[Concatenated Segments]}\\
\texttt{"""} \\
Question: \\
\blue{[Qeustion and possible Choice]} \\
If the answer CANNOT be inferred from the text above, reply with action -1.\\
If the answer CAN be inferred from the text above, reply with action -2, and provide your reasoning and the final answer. \\
You are ONLY allowed to reply with action -2 or -1. \\
\#\#\#\#\#\#\#\#\#\#\#\#\#\#\#\#\#\#\#\#\#\#\# \\
Reasoning: ... \\
Action: -2/-1, ... \\
\#\#\#\#\#\#\#\#\#\#\#\#\#\#\#\#\#\#\#\#\#\#\#  \\
\end{prompt}
\begin{prompt}[title={The prompt used for Entity Recognition}, label=prompt:entity]
\# Task Definition \\
This project aims to extract key entities from the given text to help answer related questions. Please follow the following steps when performing this task: \\
1. Understand the content: Before starting entity extraction, carefully read and understand the text's content. Pay attention to the story plot, themes, and context. \\
2. Identify key entities: Focus on important entities in the text, including but not limited to characters, locations, times, organizations, etc. These entities are usually crucial for answering questions.\\
3. Contextual relevance: When annotating entities, consider their roles and importance in the text. Note how entities are interconnected with events, plots, or other entities in the text.
4. Avoid subjective interpretation: Only annotate based on textual content and avoid adding personal interpretations or assumptions. \\
5. Review and verify: After completing the initial annotation, carefully review to ensure no omissions or errors have occurred. \\
The background information is\\
\blue{[Document Summary]}\\
\# Requirement \\
Extract the key point. Named entities from the document below are inside the triple quote and only appear in the document.\\
\texttt{"""} \\
\blue{[Segment Context]}\\
\texttt{"""} \\

\end{prompt}

\begin{prompt}[title={The prompt used for Relation Extraction}, label=prompt:relation]
\# Task Definition\\
This task aims to associate the extracted entities according to the text to help answer specific questions.\\
1. Describe the relationship in one clear paragraph;\\
2. Avoid false inferences: extract relationships based only on the clear information in the text, and avoid false inferences based on assumptions or external knowledge.\\
3. Analyze the relationships between entities: Study the mutual relationships identified by the research institute. Pay attention to how they interact and describe the relationships between entities in a concise text.\\
4. Record the context of the relationship: When extracting the relationship, record the specific context of the relationship, which is important for understanding the whole picture of the relationship.\\
\# Main text \\
\texttt{"""} \\
\blue{[Segment Content]} \\
\texttt{"""} \\
\# Extracted entities \\
\texttt{"""} \\
\blue{[Segment Content with marked entities]} \\
\texttt{"""} \\
Extract relations from the document according to the given entities, and give me the entity pair and the description for each relation.\\
\end{prompt}
\begin{prompt}[title={The prompt used for Summary}, label=prompt:summary]
\# Task Definition \\
Summary the context into shortened text, and retain as much relevant information as possible about the original characters, person, events, location, etc., in the summary. The context is \\
\texttt{"""} \\
\blue{[Segment Context]}\\
\texttt{"""} \\
\# Requirement
Summary the segments into a short paragraph.
\end{prompt}

\begin{prompt}[title={The prompt used for Relation Update}, label=prompt:update]
\# Task Definition \\
The task is to merge the corresponding relationships based on two segments, and each segment contains a corresponding relationship. In order to merge these two relationships, we may need to pay attention to some issues and generate some corresponding questions to assist in merging these two relationships, including entities that need attention, relationships, time, temporal relations, etc.

1. Ensure entities in both segments refer to the same real-world objects or concepts and compare their attributes. 

2. Verify that relationships between entities in the first segment are compatible with those in the second segment and identify any conflicts or discrepancies.

3. Check for consistent and correctly aligned time references and consider the impact of temporal relations on the merging process.

4. Refer to the document for specific guidelines or criteria for merging relationships, and review any examples or case studies that illustrate successful merging.

5. Understand how the contexts of the relationships in each segment influence the merging process and consider any external factors or conditions mentioned in the segments.

Question

\blue{[Question to be answered]}

Document

\blue{[Document summary]}

Relation \#1

Segments

\blue{[Segment \#1]}

Relations

\blue{[Relations \#1 to be merged]}

Relation \#2

Segments

\blue{[Segment \#2]}

Relations

\blue{[Relations \#2 to be merged]}
\end{prompt}

\end{document}